\documentclass{article}

\usepackage[preprint]{nips_2018}

\setcitestyle{round}	

\usepackage[utf8]{inputenc} 
\usepackage[T1]{fontenc}    
\usepackage{hyperref}       
\usepackage{url}            
\usepackage{booktabs}       
\usepackage{amsfonts}       
\usepackage{nicefrac}       
\usepackage{microtype}      

\usepackage{subfigure}

\usepackage[pdftex]{graphicx}
\usepackage{tikz}
\usetikzlibrary{positioning}
\usepackage{algorithm}
\usepackage{algorithmic}
\usepackage{amsmath}
\usepackage{amssymb}

\usepackage{soul}
\usepackage{placeins}
\usepackage{siunitx}

\usepackage{ifthen}
\newboolean{inlr}
\newboolean{midlr}
\setboolean{inlr}{false}
\newcommand\newlr{\setboolean{inlr}{false}}

\newcommand\lr[1]{%
  \ifthenelse{\boolean{inlr}}%
    {\ifthenelse{\boolean{midlr}}%
      {\middle#1\setboolean{inlr}{true}}
      {\right#1}}%
    {\left#1\setboolean{inlr}{true}}}%
\newcommand\lrmid{%
  \ifthenelse{\boolean{inlr}}%
    {\setboolean{midlr}{true}}
    {}}%

\newcommand{\angles}[2][\lr]{{\newlr #1\langle{#2}#1\rangle}}

\title{Flexible and accurate inference and learning\\ for deep generative models}

\author{
  Eszter V\'ertes \hspace{3ex} Maneesh Sahani \\
  Gatsby Computational Neuroscience Unit\\
  University College London\\
  London, W1T 4JG \\
  \texttt{\{eszter, maneesh\}@gatsby.ucl.ac.uk} \\
}

\begin{document}

\maketitle

\begin{abstract}
  We introduce a new approach to learning in hierarchical
  latent-variable generative models called the ``distributed
  distributional code Helmholtz machine'', which emphasises
  flexibility and accuracy in the inferential process.
  In common with the original Helmholtz machine and later variational
  autoencoder algorithms (but unlike adverserial methods) our approach
  learns an explicit inference or ``recognition'' model to approximate
  the posterior distribution over the latent variables.
  Unlike in these earlier methods, the posterior representation is not
  limited to a narrow tractable parameterised form (nor is it
  represented by samples).
  To train the generative and recognition models we develop an
  extended wake-sleep algorithm inspired by the original Helmholtz
  Machine.
  This makes it possible to learn hierarchical latent models with both
  discrete and continuous variables, where an accurate posterior
  representation is essential.
  We demonstrate that the new algorithm outperforms current
  state-of-the-art methods on synthetic, natural image patch and the
  MNIST data sets.
\end{abstract}

\section{Introduction}

There is substantial interest in applying variational methods to learn
complex latent variable generative models, for which the full
likelihood function (after marginalising over the latent variables)
and its gradients are intractable.
Unsupervised learning of such models has two complementary goals: to
learn a good approximation to the distribution of the observations;
and also to learn the underlying structural dependence so that the
values of latent variables may be inferred from new observations.

Variational methods rely on optimising a lower bound to the
log-likelihood (the \textit{free energy}), usually constructed by
introducing an approximation of the posterior distribution
\citep{wainwright_graphical_2008}.  The performance of variational
algorithms depends critically on the flexibility of the variational
posterior.
In cases where the approximating class does not contain the true
posterior distribution, variational learning may introduce substantial
bias to estimates of model parameters \citep{turner_two_2011}.

Variational autoencoders
\citep{rezende_stochastic_2014,kingma_auto-encoding_2014} combine the
variational inference framework with the earlier idea of the
recognition network.
This approach has made variational inference applicable to a large
class of complex generative models.
However, many challenges remain. Most current algorithms have
difficulty learning hierarchical generative models with multiple
layers of stochastic latent variables \citep{sonderby_ladder_2016}.
Arguably, this class of models is crucial for modelling data where the
underlying physical process is itself hierarchical in nature.
Furthermore, the generative models typically considered in the
literature restrict the prior distribution to a simple form, most
often a factorised Gaussian distribution, which makes it difficult to
incorporate additional generative structure such as sparsity into the
model.


We introduce a new approach to learning hierarchical generative
models, the \emph{Distributed Distributional Code (DDC) Helmholtz
  Machine}, which combines two ideas that originate in theoretical
neuroscience: the Helmholtz Machine with wake-sleep learning
\citep{dayan_helmholtz_1995}; and distributed or population codes for
distributions \citep{zemel_probabilistic_1998, sahani_doubly_2003}.
%
A key element of our method is that the approximate posterior
distribution is represented as a set of expected sufficient
statistics, rather than by directly parameterizing the probability
density function.
This allows an accurate posterior approximation without being
restricted to a rigid parametric class.
At the same time, the DDC Helmholtz Machine retains some of the
simplicity of the original Helmholtz Machine in that it does not
require propagating gradients across different layers of latent
variables.
%
%
The result is a robust method able to learn the parameters of each
layer of a hierarchical generative model with far greater accuracy
than achieved by current variational methods.

We begin by briefly reviewing variational learning
(Section~\ref{variational}), deep exponential family models (Section
\ref{deepexpfam}), and the original Helmholtz Machine and wake-sleep
algorithm (Section \ref{classichm}); before introducing the DDC
(Section \ref{ddc}), and associated Helmholtz Machine (Section
\ref{ddchm_alg}), whose performance we evaluate in Section
\ref{experiments}.

\section{Variational inference and learning in latent variable models}
\label{variational}

%
%
Consider a generative model for observations $x$, that depend on
latent variables $z$.  Variational methods rely on optimizing a lower
bound on the log-likelihood by introducing an approximate posterior
distribution $q({z}|x)$ over the latent variables:
\begin{align}
%
\log p_\theta({x}) &\geq \mathcal{F}(q, \theta, {x}) = \log p_\theta(x) - D_{KL}[q({z}|x) || p_\theta({z}|{x}) ]  \label{eq:KLqp}
%
%
%
\end{align}
The cost of computing the posterior approximation for each observation
can be efficiently amortized by using a recognition model
\citep{gershman_amortized_2014}, an explicit function (with parameters
$\phi$, often a neural network) that for each ${x}$ returns the
parameters of an estimated posterior distribution: ${x} \rightarrow
q_\phi({z}|x)$.

A major source of bias in variational learning comes from the mismatch
between the approximate and exact posterior distributions. The
variational objective penalizes this error using the `exclusive'
Kullback-Leibler divergence (see Eq.~\ref{eq:KLqp}), which
typically results in an approximation that underestimates the
posterior uncertainty \citep{minka_divergence_2005}.

Multi-sample objectives (e.g. IWAE, \citealp{burda_importance_2015}; VIMCO,
\citealp{mnih_variational_2016}) have been proposed to remedy the
disadvantages of a restrictive posterior approximation. Nonetheless,
benefits of these methods are limited in cases when the proposal
distribution is too far from the true posterior (see Section
\ref{experiments}).

%
%
%
\section{Deep exponential family models}
\label{deepexpfam}
We consider hierarchical generative models in which each conditional
belongs to an exponential family, also known as \emph{deep exponential
  family models} \citep{ranganath_deep_2015}.  Let ${x} \in
\mathcal{X}$ denote a single (vector) observation.  The distribution
of data ${x}$ is determined by a sequence of $L$ (vector) latent
variables ${z}_1 \dots {z}_L$ arranged in a conditional hierarchy as
follows:

\begin{center}
  \begin{tikzpicture}
    [xx/.style={circle,minimum size=4ex,inner sep=0pt,fill=black!20,draw=black!80},
     zz/.style={circle,minimum size=4ex,inner sep=0pt,draw=black!80}]

    \node (x) [xx] {$x$};
    \node (z1) [zz, above=3ex of x] {$z_1$} edge[->] (x);
    \node (z2) [zz, above=3ex of z1] {$z_2$} edge[->] (z1);
    \node (dots) [above=1ex of z2] {$\vdots$} edge[->] (z2);
    \node (zL) [zz, above=1ex of dots] {$z_L$} edge[->] (dots);

    \node (xeq) [right=0.5em of x] 
          {$p(x | z_1) = \exp \left( g_0(z_1,\theta_0)^T S_0(x) - \Phi_0(g_0(z_1, \theta_0))\right) $};
    \node (z1eq) [right=0.5em of z1] 
          {$p(z_1 | z_2) = \exp \left(g_1(z_2,\theta_{1})^T S_1(z_1) - \Phi_1(g_1(z_2, \theta_1))\right)$};
    \node (z2eq) [right=0.5em of z2] 
          {$p(z_2 | z_3) = \exp(g_2(z_3,\theta_{2})^T  S_2(z_2) - \Phi_2(g_2(z_3, \theta_2)))$};
    \node (zLeq) [right=0.5em of zL] 
          {$p(z_L) = \exp(\theta_L^T  S_L(z_L) - \Phi_L(\theta_L))$};
  
  \end{tikzpicture} 
\end{center}
Each conditional distribution is a member of a \emph{tractable}
exponential family, so that conditional sampling is possible.  Using
$l \in \{0, 1, 2, \dots L\}$ to denote the layer (with $l=0$ for the
observation), these distributions have sufficient statistic function
$S_l$, natural parameter given by a known function $g_l$ of both the
parent variable and a parameter vector $\theta_l$, and a log
normaliser $\Phi_l$ that depends on this natural parameter.  At the
top layer, we lose no generality by taking $g_L(\theta_L) = \theta_L$.

We will maintain the general notation here, as the method we propose
is very broadly applicable (both to continuous and discrete latent
variables), provided that the family remains tractable in the sense
that we can efficiently sample from the conditional distributions
given the natural parameters.

\section{The classic Helmholtz Machine and the wake-sleep algorithm}
\label{classichm}
The Helmholtz Machine \citep[HM;][]{dayan_helmholtz_1995} comprises a
latent-variable generative model that is to be fit to data, and a
\emph{recognition} network, trained to perform approximate inference
over the latent variables.

The latent variables of an HM generative model are arranged
hierarchically in a directed acyclic graph, with the variables in a
given layer conditionally independent of one another given the
variables in the layer above.  In the original HM, all latent and
observed variables were binary and formed a sigmoid belief network
\citep{neal_connectionist_1992} which is a special case of deep
exponential family models introduced in the previous section with
$S_l(z_l)=z_l$ and $g_l(z_{l+1},\theta_{l})=\theta_{l} z_{l+1} $.
%
%
The recognition network is a functional mapping with an analogous
hierarchical architecture that takes each ${x}$ to an estimate of the
posterior probability of each ${z}_l$, using a factorized mean-field
representation.


The training of both generative model and recognition network follows
a two-phase procedure known as \emph{wake-sleep}
\citep{hinton_wake-sleep_1995}.  In the \emph{wake} phase,
observations ${x}$ are fed through the recognition network to obtain
the posterior approximation $q_\phi({z}_l|x)$. In each layer the
latent variables are sampled independently conditioned on the samples
of the layer below according to the probabilities determined by the
recognition model parameters.
%
%
These samples are then used to update the generative parameters
to increase the expected joint likelihood -- equivalent to taking gradient steps to increase the variational free energy.
In the \emph{sleep} phase, the
current generative model is used to provide joint samples of the
latent variables and fictitious (or ``dreamt'') observations and these
are used as supervised training data to adapt the recognition network.
%
The algorithm allows for straightforward optimization since parameter updates at each layer in both the generative and recognition models are based on locally generated samples of both the input and output of the layer.

Despite the resemblance to the two-phase process of
expectation-maximisation and approximate variational methods, the
sleep phase of wake-sleep does not necessarily increase the
free-energy bound on the likelihood. Even in the limit of infinite
samples, the mean field representation $q_\phi({z}|x)$ is learnt so
that it minimises $D_{KL}[p_\theta({z}|{x}) \| q_\phi({z}|x)]$, rather
than $D_{KL}[q_\phi({z}|x) \| p_\theta({z}|{x})]$ as required by
variational learning.
For this reason, the mean-field approximation provided by the
recognition model is particularly limiting, since it not only biases
the learnt generative model (as in the variational case) but it may
also preclude convergence.


\section{The DDC Helmholtz Machine (DDC-HM)}

\subsection{Distributed Distributional Codes}
\label{ddc}
%
The key drawback of both the classic HM and most approximate variational
methods is the need for a tractably parametrised posterior
approximation.  Our contribution is to instead adopt a flexible and
powerful representation of uncertainty in terms of expected values of
large set of (possibly random) arbitrary nonlinear functions.  We call
this representation a Distributed Distributional Code (DDC) in
acknowledgement of its history in theoretical neuroscience
\citep{zemel_probabilistic_1998,sahani_doubly_2003}.
In the DDC-HM, each posterior is represented by approximate
expectations of non-linear \textit{encoding functions} $\{T^{(i)}(
z)\}_{i=1...K}$ with respect to the \emph{true posterior} $p_\theta(
z| x)$:
\begin{equation} \label{eq:recogmodel}
r_l^{(i)}({x}, \phi) \approx \angles[\big]{T^{(i)}({z}_l)}_{p_\theta({z}_l|{x})}\,,
\end{equation}
where $r_l^{(i)}({x}, \phi), i=1...K_l$ is the output of the
recognition network (parametrised by $\phi$) representing the
approximate posterior $q({z}_l|x)$ in latent layer $l=1 \dots L$ and
the angle brackets denote expectations. However, a finite---albeit
large---set of expectations does not itself fully specify the
probability distribution $p_\theta({z}|{x})$. Instead, the approximate
posterior $q({z}|x)$ is interpreted as the distribution of maximum
entropy that agrees with all of the encoded expectations.

A standard calculation shows that this distribution has a density of
the form \citep[][Ch.3]{wainwright_graphical_2008}:
\begin{equation} \label{eq:qexpfam}
  q({z}|x) = \frac1{Z(\eta({x}))} \exp\left(\sum_{i=1}^K \eta^{(i)}({x}) T^{(i)}({z})\right)
\end{equation}
where the $\eta^{(i)}$ are natural parameters (derived as Lagrange multipliers enforcing the expectation constraints), and $Z(\eta)$ is a normalising constant.  Thus, in this view, the
encoded distribution $q$ is a member of the exponential family whose
sufficient statistic functions correspond to the encoding functions  $\{T^{(i)}({z})\}$,
and the recognition network returns the expected sufficient statistics,
or \emph{mean parameters}. It follows that given a sufficiently large set of encoding functions, we can approximate the true posterior distribution arbitrarily well \citep{rahimi_uniform_2008}. Throughout the paper we will use encoding functions of the following form:
\begin{equation} \label{eq:Tsigm}
T^{(i)}({z}) = \sigma({w}^{(i)T } {z} + b^{(i)}), \textrm{ } i=1\dots K\,,
\end{equation}
where ${w}^{(i)}$ is a random linear projection with components
sampled from a standard normal distribution, $b^{(i)}$ is a similarly
distributed bias term, and $\sigma$ is a sigmoidal non-linearity. That
is, the representation is designed to capture information about the
posterior distribution along $K$ random projections in ${z}$-space.
As a special case, we can recover the approximate posterior equivalent
to the original HM if we consider linear encoding
functions $T^{(i)}({z})=z_i$, corresponding to a factorised mean-field
approximation.

Obtaining the posterior natural parameters $\{\eta^{(i)}\}$ (and thus
evaluating the density in Eq.~\ref{eq:qexpfam}) from the mean
parameters $\{r^{(i)}\}$ is not straightforward in the general case
since $Z(\eta)$ is intractable.  Thus, it is not immediately clear how
a DDC representation can be used for learning.  Our exact scheme will
be developed below, but in essence it depends on the simple
observation that most of the computations necessary for learning (and
indeed most computations involving uncertainty)
depend on the evaluation of appropriate expectations.  Given a rich
set of encoding functions $\{T^{(i)}\}_{i=1...K}$ sufficient to
approximate a desired function $f$ using linear weights
$\{\alpha_i\}$, such expectations become easy to evaluate in the DDC
representation:
\begin{align}\label{eq:fapprox}
  f({z})  \approx \sum_i \alpha^{(i)} T^{(i)}({z})  \quad 
 \Rightarrow \quad  \angles{f({z})}_{q({z})} \approx \sum_i \alpha^{(i)} \angles{T^{(i)}({z})}_{q({z})}
    = \sum_i \alpha^{(i)} r^{(i)}
\end{align}
Thus, the richer the family of DDC encoding functions, the more
accurate are both the approximated posterior distribution, \emph{and}
the approximated expectations\footnote{In a suitable limit, an infinite
  family of encoding functions would correspond to a mean embedding
  representation in a reproducing kernel Hilbert space \citep{gretton_kernel_2012}}.
We will make extensive use of this property in the following section
where we discuss how this posterior representation is learnt (sleep
phase) and how it can be used to update the generative model (wake
phase).

\subsection{The DDC Helmholtz Machine algorithm}
\label{ddchm_alg}

\begin{algorithm}
\caption{DDC Helmholtz Machine training}
\label{alg}
\begin{algorithmic}

\STATE Initialise $\theta$

\WHILE{not converged}

\STATE \textbf{Sleep phase:}
\STATE sample: ${z}_L^{(s)}, ...,  {z}_1^{(s)}, {x}^{(s)} \sim p_{\theta}({x}, {z}_1,..., {z}_L) $ 
\STATE update recognition parameters $\phi_l$ [eq.~\ref{eq:r_cost}]
\STATE update function approximators $\alpha_l, \beta_l$ [appendix]

\STATE \textbf{Wake phase:}
\STATE ${x} \leftarrow$ \{minibatch\}
\STATE evaluate ${r}_l({x}, \phi)$ [eq.~\ref{eq:r_fun}]
\STATE update $\theta$: $\Delta \theta \propto \widehat{\nabla_{\theta} {\mathcal{F}}}({x},{r}({x}, \phi),\theta)$ [appendix]
\ENDWHILE

\end{algorithmic}
\end{algorithm}

Following \citep{dayan_helmholtz_1995} the generative and recognition
models in the DDC-HM are learnt in two separate phases (see Algorithm \ref{alg}).
%
%
The sleep phase involves learning a recognition network that takes
data points ${x}$ as input and produces expectations of the non-linear
encoding functions $\{T^{(i)}\}$ as given by
Eq.~(\ref{eq:recogmodel}); \emph{and} learning how to use these
expectations to update the generative model parameters using
approximations of the form of Eq.~(\ref{eq:fapprox}).
The wake phase updates the generative parameters by computing the
approximate gradient of the free energy, using the posterior
expectations learned in the sleep phase. Below we describe the two
phases of the algorithm in more detail.

\paragraph{Sleep phase}
%
One aim of the sleep phase, given a current generative model
$p_\theta({x}, {z})$, is to update the recognition network so that the
Kullback-Leibler divergence between the true and the approximate
posterior is minimised:
\begin{equation}\label{eq:KLpq}
\phi = \mathrm{argmin}\textrm{ } \mathrm{D_{KL}}[p_\theta({z}|{x})||q_\phi({z}|x)] 
\end{equation} 
%
Since the DDC $q({z}|x)$ is in the exponential family, the
KL-divergence in Eq.~(\ref{eq:KLpq}) is minimised if the
expectations of the sufficient statistics vector ${\mathcal{T}} =
[T^{(1)}, \dots, T^{(K)} ]$ under the two distributions agree:
$ \angles[\big]{\mathcal{T} ({z})}_{p_\theta( z| x)} = \angles[\big]{\mathcal{T}({z})}_{q_\phi({z}|x)}. $ 
Hence the parameters of the recognition model should be updated so that:
${r}_l({x}, \phi_l) \approx \angles[\big]{\mathcal{T}({z}_l)}_{p_\theta( z_{l}| x)}   
$
This requirement can be translated into
an optimisation problem by sampling ${z}_L^{(s)}, \dots,
{z}_1^{(s)}, {x}^{(s)}$ from the generative model and minimising
the error between the output of the recognition model $ {r}_l({x}^{(s)}, \phi_l)$
and encoding functions $\mathcal{T}$ evaluated at the generated \emph{sleep} samples.
For tractability, we substitute the squared loss in place of Eq.~(\ref{eq:KLpq}).
%
\begin{align}\label{eq:r_cost}
\phi_l &= \mathrm{argmin} \sum_s \big({r}_l({x}^{(s)}, \phi_l) - \mathcal{T}({z}_l^{(s)})\big)^2 
%
%
\end{align}
In principle, one could use any function approximator (such as a
neural network) for the recognition model $r_l({x}^{(s)}, \phi_l)$,
provided that it is sufficiently flexible to capture the mapping from
the data to the encoded expectations.  Here, we parallel the original
HM, and use a recognition model that reflects the hierarchical
structure of the generative model. For a model with 2 layers of latent
variables:
\begin{align}\label{eq:r_fun}
{h}_1({x}, W)  &= [W {x}]_+\,, &
{r}_1({x}, \phi_1) &= \phi_1 \cdot {h}_1(W, {x})\,, &
{r}_2({x}, \phi_2) &= \phi_2 \cdot {r}_1({x}, \phi_1)\,,
\end{align}
where $W, \phi_1, \phi_2$ are matrices and $[. ]_+$ is a rectifying
non-linearity.  Throughout this paper we use a fixed $W \in
\mathsf{R}^{M \times D_x} $ sampled from a normal distribution, and
learn $\phi_1, \phi_2$ according to Eq.~(\ref{eq:r_cost}).

Recognition model learning in the DDC-HM thus parallels that of the
original HM, albeit with a much richer posterior representation.  The
second aim of the DDC-HM sleep phase is quite different: a further set
of weights must be learnt to approximate the gradients of the
generative model joint likelihood. This step is derived in the
appendix, but summarised in the following section.

\paragraph{Wake phase}
The aim in the wake phase is to update the generative parameters to
increase the variational free energy $\mathcal{F}(q, \theta)$,
evaluated on data $x$, using a gradient step:
 \begin{align}
 \Delta \theta &\propto \nabla_\theta \mathcal{F}(q,\theta)
=  \langle \nabla_\theta  \log p_\theta({x}, {z})\rangle_{q(z|x)}  \label{eq:logjoint_q}
 \end{align}
The update depends on the evaluation of an expectation over
$q({z}|x)$.
As discussed in Section \ref{ddc}, the DDC approximate posterior
representation allows us to evaluate such expectations by
approximating the relevant function using the non-linear encoding
functions $T^{(i)}$.

For deep exponential family generative models, the gradients of the
free energy take the following form (see appendix):
\begin{equation}\label{eq:gradF}
\begin{split}
\nabla_{\theta_0} \mathcal{F}
  &=  \nabla_{\theta_0} \langle \log p(x|z_1)\rangle_q  
  = S_0(x)^T \langle\nabla g(z_1, \theta_0)\rangle_{q(z_1)}
   - \langle \mu_{x|z_1}^T\nabla g(z_1, \theta_0)\rangle_{q(z_1)} \\
\nabla_{\theta_l} \mathcal{F}
  &= \nabla_{\theta_l} \langle \log p(z_l| z_{l+1}) \rangle_q  
  = \langle S_l(z_l)^T  \nabla g(z_{l+1},\theta_l)  \rangle_{q(z_{l}, z_{l+1})} 
  - \langle \mu_{z_{l}|z_{l+1}}^T \nabla g(z_{l+1}, \theta_l) \rangle_{q(z_{l+1})} \\
\nabla_{\theta_L} \mathcal{F}
  &= \nabla_{\theta_L} \langle \log p (z_L)\rangle_q 
  = \langle S_L(z_L) \rangle - \nabla \Phi(\theta_L)\,,\\[-3ex]
\end{split}
\end{equation}
where $\mu_{x|z_1}, \mu_{z_l|z_{l+1}}$ are expected sufficient
statistic vectors of the conditional distributions from the generative
model:
$\mu_{x|z_1}=\langle S_0(x) \rangle_{p(x|z_1)}$, 
$\mu_{z_l|z_{l+1}}=\langle S_l(z_l) \rangle_{p(z_l|z_{l+1})}$.
Now the functions that must be approximated are the functions of
$\{{z}_l\}$ that appear within the expectations in
Eqs.~\ref{eq:gradF}.  As shown in the appendix, the coefficients of
these combinations can be learnt by minimizing a squared error on the
sleep-phase samples, in parallel with the learning of the recognition
model.

Thus, taking the gradient in the first line of Eq.~(\ref{eq:gradF}) as
an example, we write $\nabla_\theta g({z}_1, \theta_0) \approx \sum_i
\alpha_0^{(i)} {T}^{(i)}({z}_1) = \alpha_0 \cdot T(z_l)$ and evaluate
the gradients as follows:
\begin{align}
  \text{sleep:}& & &\alpha_0 \leftarrow \mathrm{argmin} \sum_s \big(
  \nabla_\theta g(z^{(s)}_1, \theta_0) - \alpha_0 \cdot T(z_1^{(s)})\big)^2 \\
  \text{wake:}& & &\langle \nabla_\theta g({z}_1,
  \theta_0)\rangle_{q(z_1)}
    \approx \alpha_0 \cdot  \langle {T}({z}_1) \rangle_{q(z_1)}
    = \alpha_0 \cdot r_1 ({x}, \phi_1)
\end{align}
with similar expressions providing all the gradients necessary for
learning derived in the appendix.

In summary, in the DDC-HM computing the wake-phase gradients of the
free energy becomes straightforward, since the necessary expectations
are computed using approximations learnt in the sleep phase, rather
than by an explicit construction of the intractable posterior.
Furthermore, as shown in the appendix, using the function
approximations trained using the sleep samples and the posterior
representation produced by the recognition network, we can learn the
generative model parameters without needing any explicit
independence assumptions (within or across layers) about the posterior
distribution.

\section{Related work}
\label{related}
Following the Variational Autoencoder (VAE; \citep{rezende_stochastic_2014,kingma_auto-encoding_2014}), there has been a renewed interest in using recognition models -- originally introduced in the HM -- in the context of learning complex generative models. 
The Importance Weighted Autoencoder (IWAE; \citealp{burda_importance_2015}) optimises a tighter lower bound constructed by an importance sampled estimator of the log-likelihood using the recognition model as a proposal distribution.
This approach decreases the variational bias introduced by the factorised posterior approximation of the standard VAE. VIMCO \citep{mnih_variational_2016} extends this approach to discrete latent variables and yields state-of-the-art generative performance on learning sigmoid belief networks.
We compare our method to the IWAE and VIMCO in section \ref{experiments}.
\citet{sonderby_ladder_2016} demonstrate that the standard VAE has difficulty making use of multiple stochastic layers. To overcome this, they propose the Ladder Variational Autoencoder with a modified parametrisation of the recognition model that includes stochastic top-down pass through the generative model. The resulting posterior approximation is a factorised Gaussian as for the VAE.
Normalising Flows \citep{rezende_variational_2015} relax the factorised Gaussian assumption on the variational posterior. Through a series of invertible transformations, an arbitrarily complex posterior can be constructed. However, to our knowledge, they have not been successfully applied to deep hierarchical generative models.

\section{Experiments}
\label{experiments}

We have evaluated the performance of the DDC-HM on a directed
graphical model comprising two stochastic latent layers and an
observation layer. The prior on the top layer is a mixture of
Gaussians, while the conditional distributions linking the layers
below are Laplace and Gaussian, respectively:
%
\begin{align}\label{eq:ICA}
\begin{split}
p({z}_2)  &= 1/2 \left(\mathcal{N}({z}_2|m,\sigma^2)+  \mathcal{N}({z}_2|\text{--}m,\sigma^2) \right) \\
p({z}_1|{z}_2) &=  \mathrm{Laplace}({z}_1 | \mu=0,\lambda=\mathrm{softplus}(B{z}_2))  \\
p({x}|{z}_1) &=  \mathcal{N}({x}|\mu=\Lambda {z}_1, \Sigma_x=\Psi_{diag}) 
\end{split}
\end{align}
We chose a generative model with a non-Gaussian prior distribution and sparse
latent variables, models typically not considered in the VAE literature. Due to
the sparsity and non-Gaussianity, learning in these models is challenging, and
the use of a flexible posterior approximation is crucial.  We show that the
DDC-HM can provide a sufficiently rich posterior representation to
learn accurately in such a model.  We begin with low dimensional synthetic data
to evaluate the performance of the approach, before evaluating performance on a data set of natural image patches.
\begin{figure*}
\includegraphics[width=0.65\linewidth]{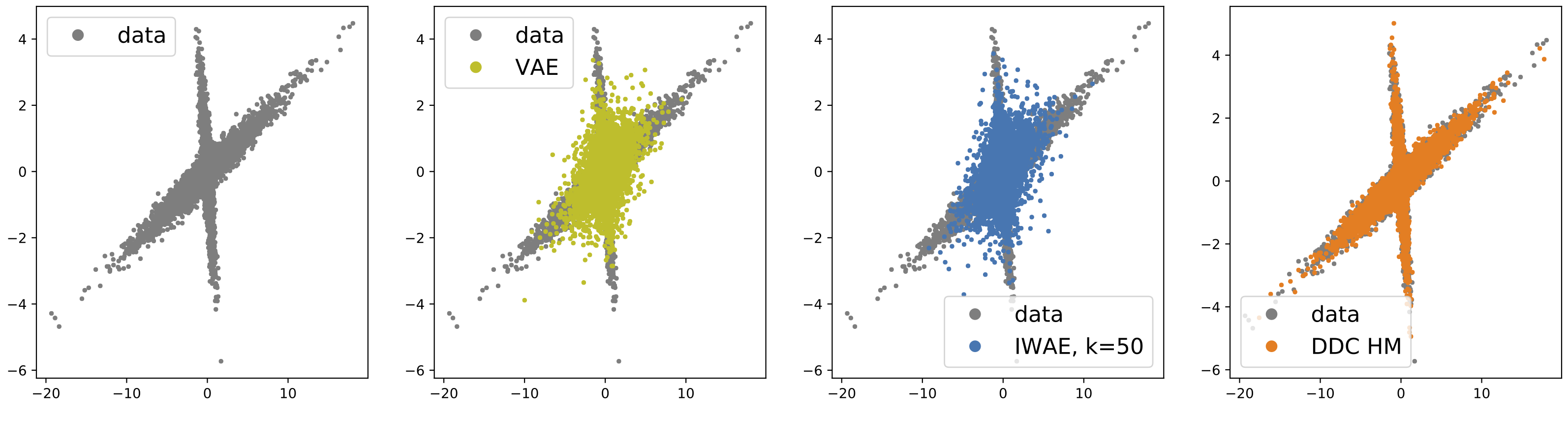}\\
\includegraphics[width=0.65\linewidth]{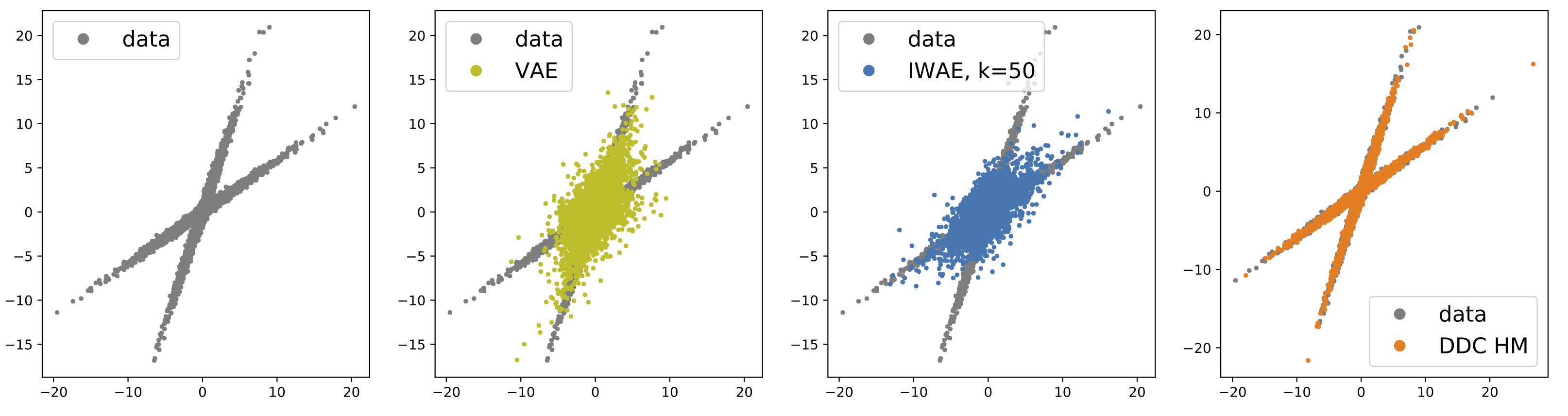}    \includegraphics[width=0.35\linewidth]{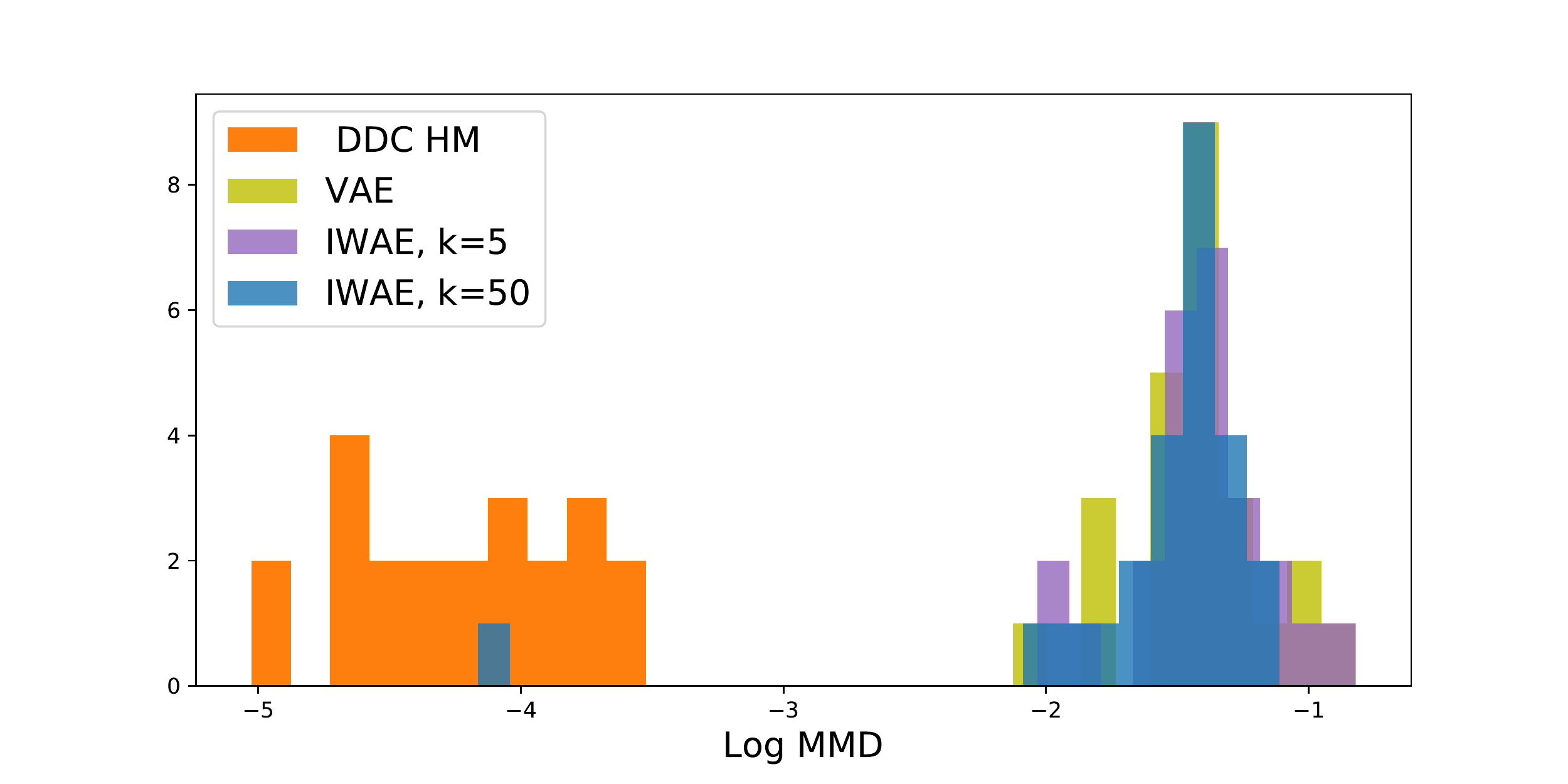}\\
\includegraphics[width=0.65\linewidth]{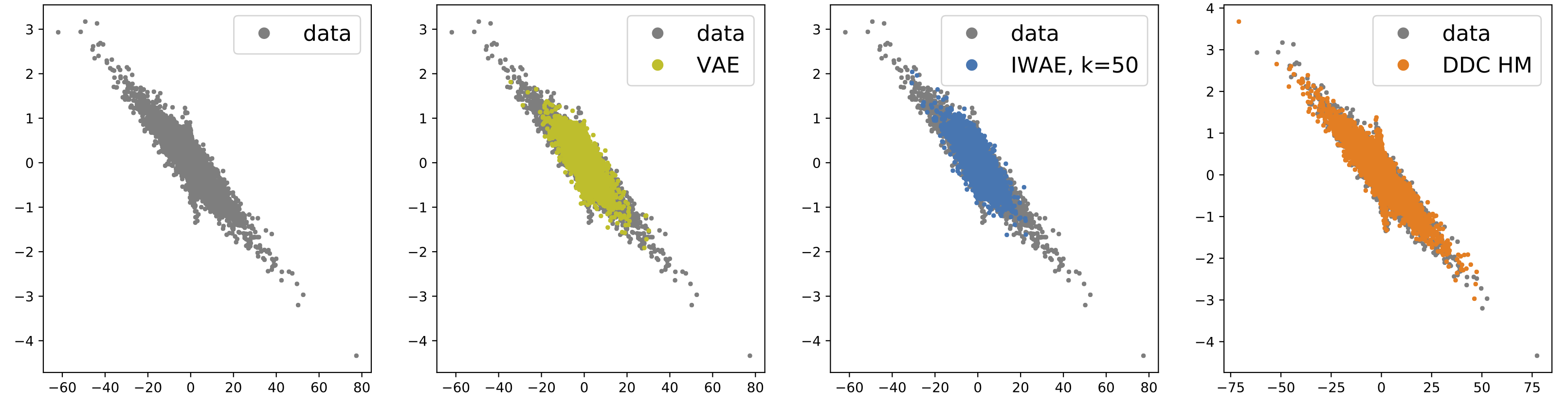}
  
  \caption{Left: Examples of the distributions learned by the Variational Autoencoder (VAE), the Importance Weighted Variational Autoencoder (IWAE) with  k=50 importance samples and the DDC Helmholtz Machine. Right: histogram of $\log$ MMD values for different algorithms trained on synthetic datasets. } \label{fig:raster2D}
\end{figure*}

\paragraph{Synthetic examples}

To illustrate that the recognition network of the DDC-HM is powerful
enough to capture dependencies implied by the generative model, we
trained it on a data set generated from the model (N=10000).  The
dimensionality of the observation layer, the first and second latent
layers was set to $D_x=2, D_1=2, D_2=1$, respectively, for both the
true generative model and the fitted models. We used a recognition
model with a hidden layer of size 100, and $K_1=K_2=100$ encoding
functions for each latent layer, with 200 sleep samples, and learned
the parameters of the conditional distributions $p(x|z_1)$ and
$p(z_1|z_2)$ while keeping the prior on $z_2$ fixed (m=3,
$\sigma$=0.1).

As a comparison, we have also fitted both a Variational Autoencoder
(VAE) and an Importance Weighted Autoencoder (IWAE), using 2-layer
recognition networks with 100 hidden units each, producing a
factorised Gaussian posterior approximation (or proposal distribution
for the IWAE). To make the comparison between the algorithms clear
(i.e. independent of initial conditions, local optima of the objective
functions) we initialised each model to the true generative parameters
and ran the algorithms until convergence (1000 epochs, learning rate:
$10^{-4}$, using the Adam optimiser; \citep{kingma_adam:_2014}).

Figure \ref{fig:raster2D} shows examples of the training data and data
generated by the VAE, IWAE and DDC-HM models after learning. The
solution found by the DDC-HM matches the training data, suggesting
that the posterior approximation was sufficiently accurate to avoid
bias during learning. The VAE, as expected from its more restrictive
posterior approximation, could capture neither the strong
anti-correlation between latent variables nor the heavy tails of the
distribution.  Similar qualitative features are seen in the IWAE
samples, suggesting that the importance weighting was unable to
recover from the strongly biased posterior proposal.

We quantified the quality of the fits by computing the \textit{maximum
  mean discrepancy} (MMD) \citep{gretton_kernel_2012} between the
training data and the samples generated by each
model \citep{bounliphone_test_2015}.
We used an exponentiated quadratic kernel with kernel width optimised
for maximum test power \citep{jitkrittum_interpretable_2016}.  We
computed the MMD for 25 data sets drawn using different generative
parameters, and found that the MMD estimates were significantly lower
for the DDC-HM than for the VAE or the IWAE (k=5, 50) (Figure
\ref{fig:raster2D}).
%
%

\begin{figure}
  \centering
  \includegraphics[width=0.24\linewidth]{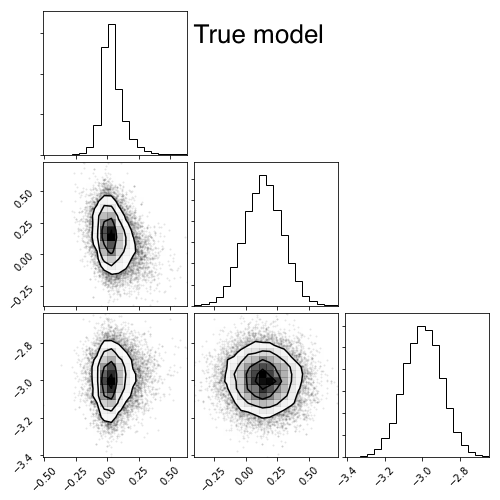}
   \includegraphics[width=0.24\linewidth]{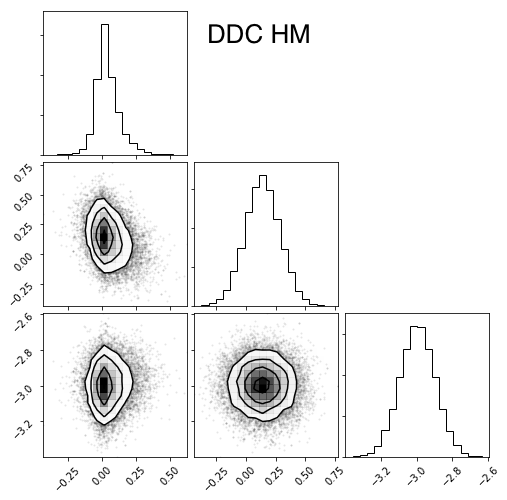} 
   \includegraphics[width=0.24\linewidth]{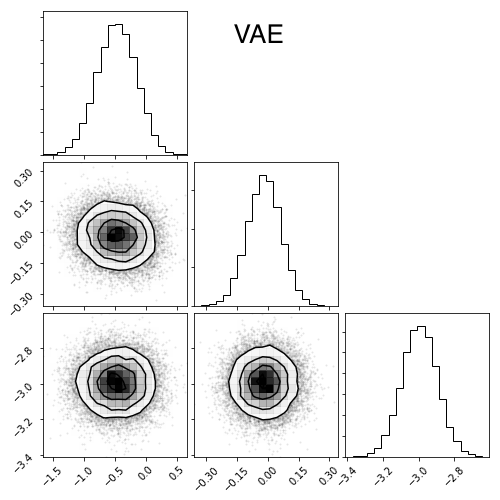}
    \includegraphics[width=0.24\linewidth]{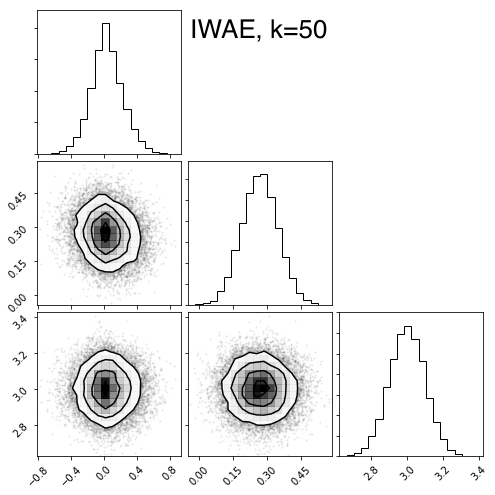}
  \caption{Example posteriors corresponding to the learned generative models. The corner plots show the pairwise and marginal densities of the three latent variables, for the true model (top left), the model learned by the VAE, IWAE (k=50) and DDC-HM.} \label{fig:posterior}
\vskip -0.2in
\end{figure}

%
Beyond capturing the density of the data, correctly identifying the underlying latent structure is also an important criterion when evaluating algorithms for learning generative models.  Figure \ref{fig:posterior} shows an example where we have used Hamiltonian Monte Carlo to generate samples from the true posterior distribution for one data point under the generative models learnt by each approach. We found that there was close agreement between the posterior distributions of the true generative model and the one learned by the DDC-HM. However, the biased recognition of the VAE and IWAE in turn biases the learnt generative parameters so that the resulting posteriors (even when computed without the recognition networks) appear closer to Gaussian.

\paragraph{Natural image patches}
We tested the scalability of the DDC-HM by applying it to a natural image data set \citep{van_hateren_independent_1998}. We trained the same generative model as before on image patches with dimensionality $D_x=16\times16$ and varying sizes of latent layers. The recognition model had a hidden layer of size $500$, $K_1 = 500$, $K_2=100$ encoding functions for $z_1$ and $z_2$, respectively, and used 1000 samples during the sleep phase.
We compared the performance of our model with the IWAE (k=50) using the relative (three sample) MMD test \citep{bounliphone_test_2015} with a exponentiated quadratic kernel (width chosen by the median heuristic). The test establishes whether the MMD distance between distributions $P_x$ and $P_y$ is significantly smaller than the distance between $P_x$ and $P_z$. We used the image data set as our reference distribution and the IWAE being closer to the data as null hypothesis. Table  \ref{mmd3} summarises the results obtained on models with different latent dimensionality, all of them strongly preferring the DDC-HM.
\begin{table}[!ht]
\caption{3-sample MMD results. The table shows the results of the `relative' MMD test between the DDC-HM and the IWAE ($k=50$) on the image patch data set for different generative model architectures. The null hypothesis tested: $\mathrm{MMD}_{\mathrm{IWAE}}<\mathrm{MMD}_{\mathrm{DDC-HM}}$. Small p values indicate that the model learned by the DDC HM matches the data significantly better than the one learned by the IWAE ($k=50$). We got similar results when comparing to IWAE $k=1,5$ (not shown).  }
\vskip 0.15in

  \label{mmd3}
  \centering
  \begin{tabular}{llllc}
    \toprule
    \multicolumn{2}{l}{\textsc{Latent dimensions}} &  IWAE &  DDC HM  & p-value \\
    \midrule
    $D_1=10$ & $D_2=10$  &  0.126 & 0.0388   & $<10^{-87}$     \\
    $D_1=50$ & $D_2=2$    & 0.0754 & 0.0269  & $<10^{-22}$       \\
    $D_1=50$ & $D_2=10 $  & 0.247 & 0.00313  &  $<10^{-141}$      \\
    $D_1=100$ & $D_2=2$  & 0.076 & 0.0211  &  $<10^{-26}$       \\
    $D_1=100$ & $D_2=10$  & 0.171 & 0.00355  &  $<10^{-89}$      \\

    \bottomrule
  \end{tabular} 
\vskip -0.1in
\end{table}

\paragraph{Sigmoid Belief Network trained on MNIST}
Finally, we have evaluated the capacity of our model to learn hierarchical generative models with discrete latent variables by training a sigmoid belief network (SBN). We used the binarised MNIST dataset of 28x28 images of handwritten images \citep{salakhutdinov_quantitative_2008}. The generative model had three layers of binary latent variables, with dimensionality of 200 in each layer. The recognition model had a sigmoidal hidden layer of size 300 and DDC representations of size 200 for each latent layer.
As a comparison, we have also trained an SBN with the same architecture using the VIMCO algorithm (as described in \citep{mnih_variational_2016}) with $50$ samples from the proposal distribution \footnote{The model achieved an estimated negative log-likelihood of 90.97 nats, similar to the one reported by \citet{mnih_variational_2016} (90.9 nats)}. To quantify the fits, we have performed the relative MMD test using the test set ($N=10000$) as a reference distribution and two sets of samples of the same size generated from the SBN trained by the VIMCO and DDC HM algorithms. Again, we used an exponentiated quadratic kernel with width chosen by the median heuristic.
The test strongly favored the DDC HM over VIMCO with $p<10^{-20}$ (with MMD values of \num{6e-4} and \num{2e-3}, respectively).

\section{Discussion}
%
%
The DDC Helmholtz Machine provides a novel way of learning hierarchical generative models that relies on a flexible posterior representation and the wake-sleep algorithm.
Due to the lack of independence or parametric assumptions of the DDC representation,
the algorithm is able to faithfully learn generative models with rich posterior distributions.

There is no need to back-propagate gradients across stochastic layers but the recognition model can be trained layer-by-layer using the samples from the generative model.
In the version discussed here, the recognition network has a restrictive form which may be updated in closed form.
However, this assumption could easily be relaxed by adding a neural network between each latent variable layer for a more powerful approximation.

Another future direction involves learning the non-linear encoding functions or choosing them in accordance with the properties of the generative model (e.g. sparsity in the random projections).
Finally, a natural extension of the DDC representation with expectations of a finite number of encoding functions, would be to approach the RKHS mean embedding, corresponding to infinitely many encoding functions \citep{grunewalder_conditional_2012}.

Even without these extensions, however, the DDC-HM offers a novel and powerful approach to probabilistic learning with complex hierarchical models.

\bibliography{journalsabbrvnodots,helmholtz_ddc} 
\newpage
\section*{\Large Appendix}

\subsection*{Computing and learning gradients for the generative model parameters}

The variational free energy for a hierarchical generative model over
observations $x$ with latent variables $z_1 \dots z_L$ can be written:
\begin{equation}
  \mathcal{F}(q, \theta)
  = \angles{\log p(x|z_1)}_{q(z_1)}
  + \sum_{l=1}^{L-1} \angles{\log p(z_l|z_{l+1}}_{q(z_l,z_{l+1})}
  + \angles{\log p(z_L)}_{q(z_L)}  + H[q(z_1 \dots z_L)]
\end{equation}
where the distributions $q$ represent components of the approximate
posterior and $H[\cdot]$ is the Shannon entropy.
We take each conditional distribution to have exponential family
form.  Thus (for example)
\begin{align}
  \log p(z_l | z_{l+1}) &= g_l(z_{l+1},\theta_l)^T S_l(z_l) - \Phi_l(g_l(z_{l+1},\theta_l))
  \intertext{from which it follows that:}
  \nabla_{\theta_l} \log p(z_l | z_{l+1})
    &= S_l(z_l)^T \nabla_{\theta_l}g_l(z_{l+1},\theta_l)
     - \Phi_l'(g_l(z_{l+1},\theta_l))
     \nabla_{\theta_l}g_l(z_{l+1},\theta_l)\\
    &= S_l(z_l)^T \nabla_{\theta_l}g_l(z_{l+1},\theta_l)
     - \mu_{z_1|z_{l+1}} \nabla_{\theta_l}g_l(z_{l+1},\theta_l)
\end{align}
where we have used the standard result that the derivative of the log
normalizer of an exponential family distribution with respect to the
natural parameter is the mean parameter
\begin{equation}
  \mu_{z_l|z_{l+1}} = \angles{S_l(z_l)}_{p_{\theta_l}(z_l | z_{l+1})} \,.
\end{equation}
Thus, it follows that:
\begin{align}
  \nabla_{\theta_0} \mathcal{F}
  &= \nabla_{\theta_0} \angles{\log
    p(x|z_1)}_{q(z_1)} = S_0(x)^T
  \langle\nabla g(z_1, \theta_0)\rangle_{q(z_1)}
  - \langle \mu_{x|z_1}^T\nabla g(z_1, \theta_0)\rangle_{q(z_1)} \label{eq:grad_bottom}\\
  \nabla_{\theta_l} \mathcal{F}
  &= \nabla_{\theta_l} \langle \log p(z_l| z_{l+1}) \rangle_q 
  = \langle S_l(z_l)^T  \nabla g(z_{l+1},\theta_l)  \rangle_{q(z_l,
    z_{l+1})} - \langle \mu_{z_l|z_{l+1}}^T \nabla g(z_{l+1},
  \theta_l) \rangle_{q(z_{l+1})}  \label{eq:joint_q}
  \intertext{(for $l = 1 \dots L-1$) and}
 \nabla_{\theta_L} \mathcal{F} & = \nabla_{\theta_L} \langle \log p (z_L)\rangle_q 
  = \langle S_L(z_L) \rangle - \nabla \Phi(\theta_L) \label{eq:grad_top} 
\end{align}
These are the gradients that must be computed in the wake phase.

In order to compute these gradients from the DDC posterior
representation we need to express the functions of $z_l$ that appear
in above as linear combinations of the encoding functions $T(z_l)$.
The linear coefficients can be learnt using samples from the
generative model produced during the sleep phase.
The example given in the main paper is most straightforward.  We wish
to find
\begin{align}
  \nabla_{\theta_0} g({z}_1, \theta_0) &\approx \sum_i \alpha_0^{(i)} {T}^{(i)}({z}_1)
\end{align}
in which the coefficients $\alpha_0$ can be obtained from
samples by evaluating the gradient of $g$ with respect to $\theta_0$
at $z_1^{(s)}$ and minimising the squared error:
\begin{align}
  \alpha_0 &\leftarrow \mathrm{argmin} \sum_s \big(\nabla_\theta g(z^{(s)}_1, \theta_0) - \alpha_0 \cdot T(z_1^{(s)})\big)^2\,.
\end{align}
Once these coefficients have been found in the sleep phase, the wake
phase expectations are found from the DDC recognition model very simply:
\begin{align}
  \langle \nabla_{\theta_0} g({z}_1, \theta_0)\rangle_{q(z_1)} & \approx \sum_i \alpha^{(i)}  \langle {T}^{(i)}({z}_1) \rangle_{q(z_1)} \\
  & = \sum_i \alpha^{(i)} r_1^{(i)} ({x}, \phi_1)
 \end{align}

Some of the gradients (see Eq.\ref{eq:joint_q}) require taking
expectations using the joint posterior distribution $q({z}_l,
{z}_{l+1}|{x})$. However, the recognition network as we have described
it in the main paper only contains information about the marginal
posteriors $q({z}_l| {x}),q({z}_{l+1} |{x})$.  It turns out that it is
nevertheless possible to estimate these expectations without imposing
an assumption that the approximate posterior factorises across the
layers ${z}_l, {z_{l+1}}$.

We begin by noticing that due to the structure of the generative model
the posterior distribution $q$ can be factorised into $q({z}_{l+1},
{z}_l| {x})=p({z}_{l+1}| {z}_l) q({z}_l| {x})$ without any further
assumptions (where $p({z}_{l+1}| {z}_l) $) is the conditional implied
by the generative model. Thus, we can rewrite the term in Equation
\ref{eq:joint_q} as:
$$ \langle S_l(z_l)^T \langle \nabla g(z_{l+1},\theta_l) \rangle_{p(z_{l+1}|z_l)} \rangle_{q(z_l)}$$
Now, we can replace the expectation $\langle \nabla
g(z_{l+1},\theta_l) \rangle_{p(z_{l+1}|z_l)}$ by an estimate using
\emph{sleep} samples, i.e. for each pair of samples $\{{z}_l^{(s)},
     {z}_{l+1}^{(s)}\}$ from the prior, we have a single sample from
     the true posterior distribution $ {z}_{l+1}^{(s)} \sim
     p({z}_{l+1}|{z}_l= {z}_l^{(s)}) $.
Thus, we obtain coefficients $\alpha_l$ during the sleep phase so:
\begin{align}
  \alpha_l
    &= \text{argmin} \sum_s \big( S_l(z^{(s)}_l)^T \nabla_{\theta_l}
  g(z^{(s)}_{l+1}, \theta_l) - \alpha_l^T T(z^{(s)}_l) \big)^2
\end{align}
which will converge so that
\begin{align}
  \alpha_l^T T(z_l) \approx S_l(z_l)^T \angles{\nabla_{\theta_l} g(z_{l+1}, \theta_l)}_{p(z_{l+1}|z_l)}\,.
\end{align}
Now, during the wake phase it follows that
\begin{align}
  \angles{S_l(z_l)^T \nabla g(z_{l+1},\theta_l) }_{q(z_{l+1}, z_l)}
  &=  \langle S_l(z_l)^T \langle \nabla g(z_{l+1},\theta_l) \rangle_{p(z_{l+1}|z_l)} \rangle_{q(z_l)}\\
  & \approx \angles{\sum_i \alpha_l^{(i)}  {T}^{(i)}({z}_1)}_{q(z_1)} \\
  & = \sum_i \alpha_l^{(i)} r_l^{(i)} ({x}, \phi_l)
\end{align}

Thus, all the function approximations needed to evaluate the gradients
are carried out by using samples from the generative model
(\emph{sleep} samples) as training data.
The full set of necessary function approximations with (matrix)
parameters describing the linear mappings denoted by
$\{\alpha_l\}_{l=0 \dots L}, \{\beta_l\}_{l=1\dots L}$ is:
\begin{align}
\alpha_0: T(z_l^{(s)}) &\rightarrow \nabla_{\theta}g(z_l^{(s)},\theta) \label{eq:delta0} \\
%
%
\beta_l: T(z_l^{(s)}) &\rightarrow \mu_{z_{l-1}|z_l^{(s)}}^T \nabla_{\theta} g(z_l^{(s)}, \theta)  \label{eq:gammal} \\
\alpha_l: T(z_l^{(s)}) &\rightarrow  S_l(z_l^{(s)})  \nabla_{\theta} g(z_{l+1}^{(s)}, \theta)  \label{eq:delta} \\
\alpha_L: T(z_L^{(s)}) &\rightarrow S_L(z_L^{(s)}) \label{eq:alpha}
\end{align}
where the expressions on the right hand side are easy to compute given
the current parameters of the generative model. Note that $
\mu_{z_{l-1}|z_l^{(s)}}$ appearing in equation \ref{eq:gammal} are
expectations that can be evaluated analytically for tractable
exponential family models, as a consequence of the conditionally
independent structure of the generative model. Alternatively, they can
also be estimated using the \emph{sleep} samples, by training
\begin{align}
\beta_l: T(z_l^{(s)}) &\rightarrow {S_{l-1}(z^{(s)}_{l-1})}^T \nabla_{\theta} g(z_l^{(s)}, \theta)
\end{align}

Finaly, putting the sleep and the wake phase together, the updates for
the generative parameters during the wake phase are:
\begin{equation}\label{eq:wake_gradients}
\begin{split}
\Delta \theta_0  &\propto S_0({x})^T \alpha_0 r_1({x}, \phi_1) - \beta_1 r_1({x},\phi_1) \\
\Delta \theta_{l} &\propto \alpha_l r_l({x}, \phi_l) - \beta_{l+1} r_{l+1}({x},\phi_{l+1})  \\
\Delta \theta_L &\propto \alpha_L r_L({x}, \phi_L) - \nabla_{\theta_L} \Phi(\theta_L) 
\end{split}
\end{equation}

\end{document}